# Person Re-identification Based on Color Histogram and Spatial Configuration of Dominant Color Regions

Kwangchol Jang, Sokmin Han, Insong Kim

College of Computer Science, **KIM IL SUNG** University, Pyongyang, D.P.R of Korea

*Abstract* – There is a requirement to determine whether a given person of interest has already been observed over a network of cameras in video surveillance systems. A human appearance obtained in one camera is usually different from the ones obtained in another camera due to difference in illumination, pose and viewpoint, camera parameters. Being related to appearance-based approaches for person re-identification, we propose a novel method based on the dominant color histogram and spatial configuration of dominant color regions on human body parts. Dominant color histogram and spatial configuration of the dominant color regions based on dominant color descriptor(DCD) can be considered to be robust to illumination and pose, viewpoint changes. The proposed method is evaluated using benchmark video datasets. Experimental results using the cumulative matching characteristic(CMC) curve demonstrate the effectiveness of our approach for person re-identification.

## 1. Introduction

Person re-identification is to recognize the persons previously observed by many cameras, and is to identify individuals among many candidates. Person re-identification needs techniques of matching features between individual from a probe set and the corresponding template in a gallery set.

Depending on the number of available frames per individual, the scenarios for person re-identification can be classified by Single vs Single(SvsS), Multiple vs Single(MvsS), Multiple vs Multiple(MvsM)[3].

On the other hand, the recent person re-identification approaches can be divided into non-learning based methods, and learning-based methods. And also human body can be subdivided with respect to its symmetry properties. Anti-symmetry subdivides into head, torso and legs, while symmetry can be used to divide into left and right parts of human body.

In [1], human body part detector is based on fifteen non-overlapping square cells of individual body. And also in [2], firstly Haar-like features are extracted from the full body, and then the body is divided into upper and lower part, each is described by the MPEG-7 Dominant Color Descriptor. In [4], using interest points and Hessian-Affine interest operator is proposed, and also AdaBoost-based learning method [5], a matching distance learning [8] is presented. Also using Global Color Context(GCC) [6], MPEG-7 Dominant Color Descriptor(DCD)[12,13,14], Maximal Stable Color Region(MSCR) [9,10,11] and so on are proposed.

Our person re-identification approach is based on Dominant Color Descriptor(DCD), that is, Dominant Color Regions of human body parts. The rest of the paper is organized as follows. Section 2 points out the extractions of Dominant Color Descriptor and Dominant Color Region, and section 3 addresses the feature matching and a framework for our proposed method. Experimental results and conclusions are given in section 4 and 5, respectively.

## 2. Dominant Color Descriptor and Dominant Color Region Extraction

### 2.1. Color Space Choice and Quantization

Color feature is important in image processing and recognition, video surveillance, and so on. Especially, for appearance-based person re-identification, color is the most expressive and powerful cue. The HSV color space provides an intuitive representation and approximates the way in which human perceives and manipulates the color. This color model is already represented by Munsell 3D space reference frame. In general, the color value can be transformed conveniently from RGB to HSV [13].

The conversion algorithm is shown in follow:

$C_{RGB}=(R,G,B)$ is a color value of RGB color space,

$C_{HSV}=(h,s,v)$ is the transformed color value of HSV color pace.

$r=R/255$,  $g=G/255$,  $b=B/255$   ($h,s,v \in [0,1]$)

$MAX=\max(r,g,b)$, $MIN=\min(r,g,b)$

$v=MAX$, $delta=MAX-MIN$

if ($MAX==0$)

$\quad\quad s = 0$ ;

else $s=delta/Max$;

if ($MAX==MIN$)   $h=-1$;

//Hue is undefined. (Achromatic color)

else

{

if ( $r==MAX$ && $g \neq MIN$ )

$h = 60*(g-b)/delta$;

else if ($r==MAX$ && $g==MIN$)

$\quad h = 360+60*(g-b)/delta$;

else if ($g==MAX$)

$\quad h = 60*(2.0+ (b-r)/delta)$;

else

$\quad h=60*(4.0+(r-g)/delta)$;

}

where for a color value ($h,s,v$) of HSV color space, $h \in [0, 360]$ , $s \in [0,1]$ , $v \in [0,1]$.

The dimension of the color histogram of HSV color space used to describe the color features directly, will be very much for true-color images especially, owing to the abundance of image color information. Therefore, it is essential to reduce the dimension of HSV color space components. For HSV color space, we divide hue $H$ into eight parts, saturation $S$ and intensity $V$ into three parts respectively in consideration of the human eyes to distinguish.

A non-interval quantization algorithm of the HSV space is shown as follows [13].

$$H = \begin{cases} 0 & if\ h \in [316,20) \\ 1 & if\ h \in [20,40) \\ 2 & if\ h \in [40,75) \\ 3 & if\ h \in [75,155) \\ 4 & if\ h \in [155,190) \\ 5 & if\ h \in [190,270) \\ 6 & if\ h \in [270,295) \\ 7 & if\ h \in [295,316) \end{cases} \quad (1)$$

$$S = \begin{cases} 0 & if\ s \in [0,0.2] \\ 1 & if\ s \in (0.2,0.7] \\ 2 & if\ s \in (0.7,1] \end{cases} \quad (2)$$

$$V = \begin{cases} 0 & if\ v \in [0,0.2] \\ 1 & if\ v \in (0.2,0.7] \\ 2 & if\ v \in (0.7,1] \end{cases} \quad (3)$$

where $h \in [0,360]$, $s \in [0,1]$, and $v \in [0,1]$.

According to the quantization levels as above, the $H,S,V$ 3-dimensional feature vector for different pixels can be transformed into 1-dimensional feature vector $C$ as follows:

$$C = Q_S Q_V H + Q_V S + V \quad (4)$$

where $Q_S$, $Q_V$ are quantified series of $S$ and $V$, respectively, then $Q_S = Q_V = 3$.

$$C = 9H + 3S + V \quad (5)$$

Such value $C$ is called quantified value of pixel. And all the number of possible quantified values is 8*3*3=72, and also quantified value $C$ of pixel belongs to [0,71]. 0,1,···,71 are called quantization levels for 1-dimensional feature vector, and $L$=72 is called the number of quantization labels. For HSV color feature vector (0,0,0), just $C$=0, and for HSV color feature vector (7,2,2), $C$=9*7+3*2+2=71.

As above, RGB 3-D color feature vector of each pixel is just transformed into 1-D quantified feature vector through RGB to HSV transformation, and it is limited to $L$ levels. Such quantification can be effective in reducing the dimension by effects of light intensity, but also reducing the computational time and complexity.

**2.2. Dominant Color Descriptor Extraction**

Dominant Color Descriptor(DCD) is defined as $F = \{\{C_i, P_i, V_i\}, S\}$, $i$=1,2,···,$N$ [13]. Here $C_i$ is $i$th dominant color, $P_i$ is the percentage for dominant color $C_i$, $V_i$ is its color variance, and also $S$ is a single number that represents the overall spatial homogeneity of the dominant color in the image, and $N$ is the dominant color's number. Now without consideration of color variance $V$ and dominant color spatial uniformity $S$, the above expression is simplified as $F = \{C_i, P_i\}$, $P_i \in [0,1]$, $i$=1,2,···,$N$. Also, based on the above quantification, $N=L$=72.

On the other hand, the number of dominant colors can vary from image to image and a maximum of eight dominant colors can be used to represent the regions in MPEG-7 [13,15].

Fig.1 shows DCD representation of regions. Here each region can be represented by two to four dominant colors. Thus a few dominant colors are enough to represent a region. In fact, MPEG-7 recommends that a region may have one to eight dominant colors, and a maximum of eight dominant colors is extracted for a region.

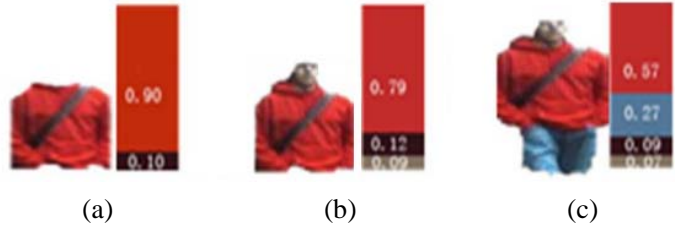

(a)　　　　　　(b)　　　　　　(c)

Fig.1. DCD representation of regions.　Regions in (a), (b) and (c) need 2, 3 and 4 centroid dominant colors.

Such eight dominant colors are called centroid dominant colors, each descriptor of them is called centroid Dominant Color Descriptor(simplicity centroid DCD). And also, let $M$ be the maximum number of centroid dominant colors, $M=8$. For a given image, $M$ centroid DCDs extraction algorithm is shown as follows.

**Input:**　　$I$: given image, $L$: quantization level number ($L=72$)
**Output:**　$F^I$: set of $M$ centroid DCDs of image $I$

　　$F^I=\varnothing$;

Step 1. for all pixels of $I$, calculate the quantified value $C$. $C=9*H+3*S+V$.

Step 2. for $i=0,1,\cdots,L-1$, calculate the normalized histogram $P_i$ of color $i$.

　　$F^I=F^I \cup \{F_i=\{i,P_i\}\}$;

Step 3. for $F_i \in F^I$　($i=0,\cdots,L-1$) , rearrange $F^I$ in descending order of $P_i$.

Step 4. Leave only first $M$ dominant color descriptors as centroid DCDs in $F^I$.

Step 5. for $F_i=\{C_i,P_i\} \in F^I$, renormalize $P_i$. ($i=1,2,\cdots,M$)

### 2.3. Extraction of Dominant Color Regions

For image $I$ and its $M$ centroid DCDs $F^I$, let $C^I$ be a set of centroid dominant colors of image $I$.

$$C^I=\{C_i| \{C_i,P_i\} \in F^I\}, i=1,\cdots,M$$

where $C_i$ is $i$th centroid dominant color.

A dominant color region of image $I$ is a connected component of pixels with its corresponding dominant color.

Therefore, for image $I$, the extraction of dominant color regions can be just considered being similar to the connected component extraction of a grey image.

The dominant color region extraction algorithm is shown as follows:

**Input:** $I$: given image, $C^I$: set of $M$ centroid dominant colors of $I$
**Output:** $R$: set of dominant color regions

$R=\varnothing$;

For all the centroid dominant colors, $C_i \in C^I$, $i=1,2,\cdots,M$, repeat the following steps 1 to 3;

Step 1. extract the set of pixels with dominant color $C_i$, $I_{C_i}$.

Step 2. for $I_{C_i}$, extract the set of all the connected components $R_{C_i} = \{R_1^{C_i}, R_2^{C_i}, ..., R_{N_i}^{C_i}\}$.

Here, $N_i$ is a number of the connected components with dominant color $C_i$, $R_k^{C_i}$ is the $k$th connected component with the dominant color $C_i$, $R_k^{C_i} \subset I_{C_i}$. ($k \in [1, N_i]$)

Step 3. $R = R \cup R_{C_i}$.

where $R$ is the set of segmented dominant color regions, and each region of $R$ has the corresponding centroid DCD. And the relatively small regions with 1 to 4 pixels can be regarded as noise regions, and should be removed. To do this, the thining algorithm is applied to each region of $R$ extracted as above, and then the resulting empty regions with no pixel are removed from $R$, just regarding as noise region, and the relatively large regions are left in $R$ as before.

The extraction of dominant color regions is shown in Fig.2.

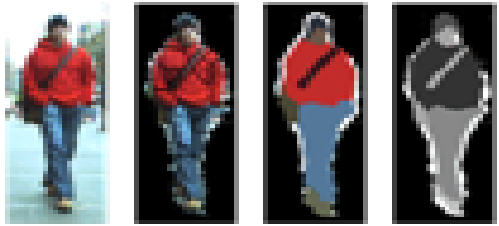

Fig.2. Extraction of DCRs.

## 3. Feature Matching

In general, for person re-identification, a probe person pedestrian image $A$ and gallery person pedestrian image set $G_B$ are given. Person re-identification is to retrieve the candidate pedestrian images of $G_B$ similar to probe pedestrian image $A$ and rearrange them in similarity order. Here, it considers feature matching between two images: a probe pedestrian image $A$ and a pedestrian image $B$ of gallery set $G_B$.

It assumes that human full body is extracted from given image by foreground and background segmentation beforehand. And then human full body is divided to only two body parts: upper and lower body parts by anti-symmetry as presented in Farenzena et al. [3]. The extraction of the dominant color descriptors and dominant color regions is applied to human body parts.

Consequently, the feature matching for person re-identification is performed between the dominant color regions extracted from human body parts of two images *A* and *B*.

### 3.1. Matching Based on Dominant Color Histogram Similarity of Human Body Parts

In general, the color cue is widely used in person re-identification as well as image retrieval and so on. Especially, the color of clothing provides the considerable information to identify the individuals. The histogram of clothing color is just a key feature widely used in many applications for person re-identification, particularly appearance-based approach.

Peoples wear the clothes with similar or equal colors in upper and lower body parts: torso and legs, while wear the clothes with different colors. In accordance with clothing of human body parts: torso and legs, the color histograms of each part can be similar or different. Here, notice that color for histogram is regarded as dominant color.

Fig.3 is shown the color histogram similarity according to clothing. Fig.3(a) is depicted that color histograms of two full bodies are similar, while both color histograms of upper parts and ones of lower parts, are different respectively. However, Fig.3(b) is presented that color histograms of two full bodies are similar, and ones of upper parts and lower parts are similar too.

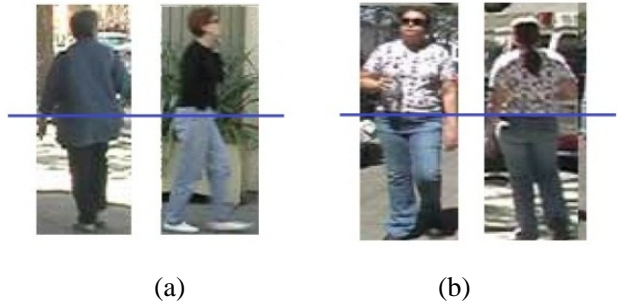

(a)          (b)

Fig.3. Histogram similarity of different images.
(a). similar for full body, while different for upper and lower parts, respectively.
(b). similar for full body, and also similar for upper and lower parts, respectively.

Let $A_U$, $A_L$ be sets of pixels in upper and lower part of probe image *A*, respectively. And also similarly let $B_U$, $B_L$ be sets of pixels in upper and lower part of gallery image *B*, respectively.

And let $F_{A_U}$, $F_{A_L}$ be sets of $M$ centroid DCDs of $A_U$ and $A_L$, respectively. Also let $F_{B_U}$, $F_{B_L}$ be sets of $M$ centroid DCDs of $B_U$ and $B_L$ respectively. That is,

$$F_{A_U} = \{\{C_i^{A_U}, P_i^{A_U}\} | i = 1, 2, \cdots, M\}$$

$$F_{A_L} = \{\{C_i^{A_L}, P_i^{A_L}\} | i = 1, 2, \cdots, M\}$$

$$F_{B_U} = \{\{C_i^{B_U}, P_i^{B_U}\} | i = 1, 2, \cdots, M\}$$

$$F_{B_L} = \{\{C_i^{B_L}, P_i^{B_L}\} | i = 1, 2, \cdots, M\}$$

where $M=8$, and also $P_i^{A_U}$, $P_i^{A_L}$, $P_i^{B_U}$, and $P_i^{B_L}$ are the percentages of $i$th quantified dominant color $C_i^{A_U}$, $C_i^{A_L}$, $C_i^{B_U}$, $C_i^{B_L}$ in $A_U$, $A_L$, $B_U$, $B_L$, respectively, and also are normalized. Similar to histogram intersection, the color histogram similarity $d_{\text{DCH}}(A,B)$ between two images $A$ and $B$, is defined as follows:

$$d_{\text{DCH}}(A,B) = \gamma d(A_U, B_U) + (1-\gamma)d(A_L, B_L) \tag{6}$$

$$d(A_U, B_U) = \sum_{i=0}^{71} \min(P_i^{A_U}, P_i^{B_U}) \tag{7}$$

$$d(A_L, B_L) = \sum_{i=0}^{71} \min(P_i^{A_L}, P_i^{B_L}) \tag{8}$$

where $d(A_U,B_U)$ is the color histogram similarity between upper body parts of two images $A$ and $B$, and also $d(A_L,B_L)$ is the color histogram similarity between lower body parts of $A$ and $B$. Both $d(A_U,B_U)$ and $d(A_L,B_L)$ are normalized. Here, $\gamma$ is a weighted coefficient, $0<\gamma<1$.

$d_{\text{DCH}}(A,B)$ is just the weighted sum of color histogram similarities $d(A_U,B_U)$ and $d(A_L,B_L)$, and is also normalized. The larger $d(A_U,B_U)$ is, the more similar upper body parts of two images $A$ and $B$ are. And also the larger $d(A_L,B_L)$ is, the more similar lower body parts of two images $A$ and $B$ are.

Consequently, the larger $d_{\text{DCH}}(A,B)$ is, the more similar two images $A$ and $B$ are deemed to be.

$d_{\text{DCH}}(A,B)$ can be regarded as color histogram similarity, related to statistical dominant color configuration of human clothing.

### 3.2. Matching Based on Spatial Similarity of Dominant Color Regions

Let $R_A$, $R_B$ be sets of dominant color regions of images $A$ and $B$, respectively.

$$R_A = \{r_i^A\}, i=1,2,\cdots,N_A,$$

$$R_B = \{r_j^B\}, j=1,2,\cdots,N_B$$

Being based on dissimilarity, the spatial similarity $d_{\text{DCR}}(A,B)$ based on dominant color region between two images $A$ and $B$, is defined by dissimilarity as follows:

$$d_{\text{DCR}}(A,B) = \sum_{u \in R_A} \min_{w \in R_B, u_c = w_c} d_R(u,w) \tag{9}$$

where $u$ and $w$ are dominant color regions of $R_A$ and $R_B$, respectively. $u_c$ and $w_c$ are dominant colors of $u$ and $w$, respectively. $d_R(u, w)$ is defined by the dissimilarity between two regions $u$ and $w$ as follows:

$$d_R(u, w) = \beta d_y(u, w) + (1-\beta)d_h(u, w) \tag{10}$$

$$d_y(u, w) = |u_y - w_y|/H \tag{11}$$

$$d_h(u, w) = |u_h - w_h|/H \tag{12}$$

where $u_y$, $w_y$ are $y$ components of center of minimum bounding rectangles(MBRs) for region $u$ and $w$, respectively. Similarly, $u_h$, $w_h$ are heights of MBRs for region $u$ and $w$, respectively. $H$ is the normalized height of two images $A$ and $B$. In fact, the sizes of two test images are normalized to be equal beforehand. Here, for each region's MBR, $x$ component of center and its width are not considered, being not robust to pose and viewpoint changes.

As shown in (11), (12), $d_y(u, w)$ is the distance between the centers of MBRs for $u$ and $w$, and $d_h(u, w)$ is the difference between the heights of MBRs for $u$ and $w$. Both $d_y(u, w)$ and $d_h(u, w)$ are normalized. Here, $\beta$ is the weighted coefficient, $0<\beta<1$, therefore $d_R(u, w)$ is normalized too. If two dominant color regions $u$ and $w$, are closed on $y$ axis and heights of MBRs are similar, $d_R(u, w)$ is small approximately to be zero, consequently it can be regarded that they are deemed to be similar.

The smaller $d_{\text{DCR}}(A,B)$ is, the more similar $A$ and $B$ are deemed to be. $d_{\text{DCR}}(A,B)$ is just regarded as a similarity based on the spatial configuration of dominant color regions related to the person clothing.

### 3.3. Integrated Feature Matching

Similarity between probe image $A$ and gallery image $B$, is evaluated as follows:

$$d(A,B)=\alpha d_{\text{DCH}}(A,B)+(1-\alpha)d_{\text{DCR}}(A,B) \tag{13}$$

where $d(A,B)$ is the similarity between two images $A$ and $B$. Here, $\alpha$ is a weighted coefficient, $0<\alpha<1$, $d(A,B)$ is normalized.

If for two images $A$ and $B$, the dominant color histograms between upper parts are similar and also the dominant color histograms between lower parts are similar, and also the spatial configuration of dominant color regions of them is similar, $d(A,B)$ is relatively larger, and also it can be regarded that they are deemed to be similar.

Through some experiments, the parameters for $d(A,B)$ are set to as follows: $\alpha=0.4<1$, $\beta=0.6<1$, $\gamma=0.55<1$. Here, $\alpha=0.4<1$ is that spatial configuration feature $d_{\text{DCR}}$ is a little more important than color configuration feature $d_{\text{DCH}}$. $\beta=0.6<1$ is that $d_y$ denoted the closeness of dominant color regions is a little more important than $d_h$ related to the heights of them,. And also, $\gamma=0.55<1$ is that clothing color configuration of upper part: torso, is more complicated than ones of lower part: legs, and therefore $d(A_U,B_U)$ is a little more significant than $d(A_L,B_L)$.

In Fig.4, the framework of our approach is shown.

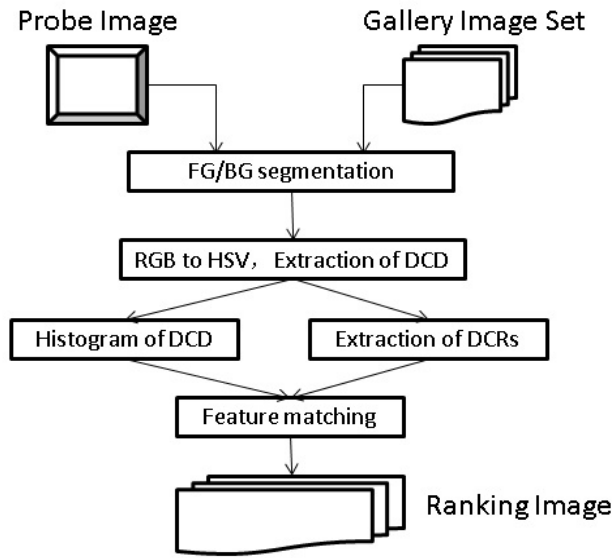

Fig.4. Framework of our approach.

## 4. Experimental Results

To evaluate the performance of our approach, we conducted some experiments with the highly challenging VIPeR dataset(single-shot scenario), being one of the benchmark datasets. The person re-identification performance can be presented using the Cumulative Matching Characteristic(CMC) curve.

The VIPeR dataset consists of 632 person image pairs by two different camera views. Fig 5 shows the CMCs of our approach and others；ELF，SDALF，and ERSVM. And also some of results: query images and target images, are shown in Fig 6. Here, query images 1, 2, and 4 matched correctly in first rank order, and query image 3 didn't.

The result of CMCs presents that our approach is more effective than other methods as above.

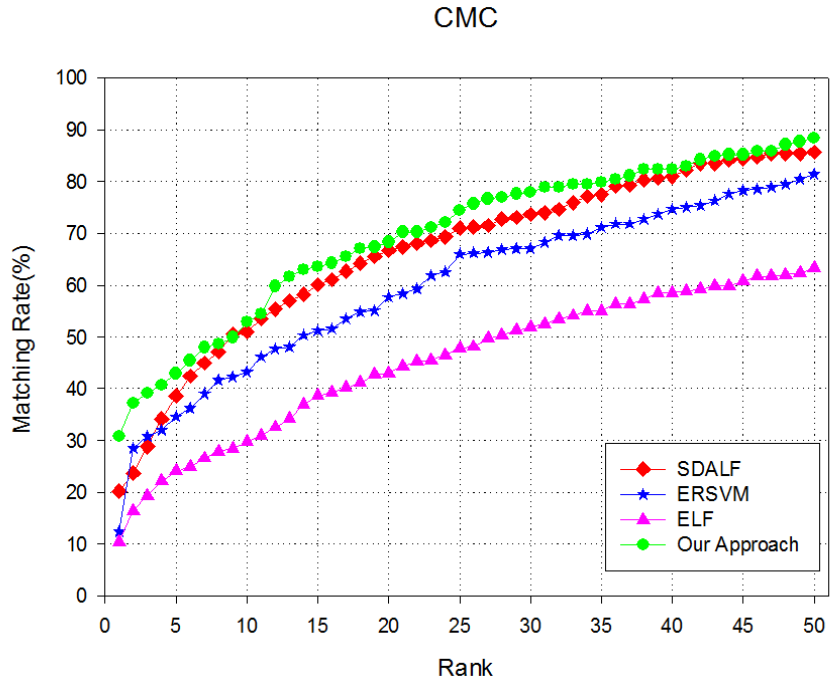

Fig.5. CMCs of Our Approach and other methods for VIPeR dataset.

## 5. Conclusions

In this paper, we addressed the person re-identification based on the color and spatial configurations of dominant color regions of human body parts. Proposed approach is a kind of appearance-based person re-identification.

It is just based on the dominant color histograms and the spatial distributions of dominant color regions of clothing. Dominant color histogram similarity is related to color configuration, and also similarity of dominant color regions is related to spatial configuration for our appearance-based approach.

The experiment results presented that our approach should be robust to pose, viewpoint and illumination changes, and also more effective than others.

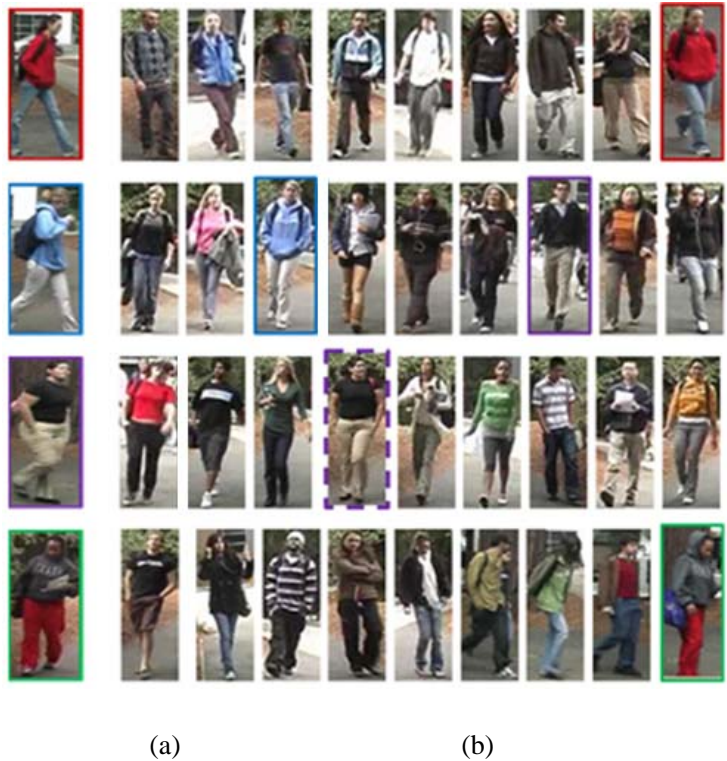

(a) (b)

Fig.6. Results. (a) query images. (b) target images.